\documentclass[pdflatex,sn-mathphys-num]{sn-jnl}
\usepackage{graphicx}
\usepackage{xcolor}
\let\orcidlogo\relax
\usepackage{orcidlink}
\hypersetup{hidelinks,bookmarksdepth=2}

\usepackage{amsmath,amssymb}
\usepackage{array}
\usepackage{booktabs}
\usepackage{makecell}
\usepackage{needspace}
\usepackage{etoolbox}
\let\unpatchedsection\section
\pretocmd{\section}{\Needspace{6\baselineskip}}{}{}
\pretocmd{\subsection}{\Needspace{7\baselineskip}}{}{}
\pretocmd{\subsubsection}{\Needspace{6\baselineskip}}{}{}
\newcommand{\code}[1]{{\def\{{\textnormal{\textbraceleft}}\def\}{\textnormal{\textbraceright}}\texttt{#1}}}
\begin{document}
\title[Correctness Boundaries of Typed Fleet Plans]{Correctness Boundaries of Typed Plans for Natural-Language Robot Fleet Control}
\author*[1]{\fnm{Artem}\sur{Tuzov}\hspace{2pt}\orcidlink{0009-0007-5252-1886}}\email{a.tuzov@innopolis.university}
\author[1]{\fnm{Dmitrii}\sur{Vizitei}\hspace{2pt}\orcidlink{0009-0006-6275-0589}}
\author[1]{\fnm{Evgenii}\sur{Shlomov}\hspace{2pt}\orcidlink{0009-0005-8864-5819}}
\author[1]{\fnm{Senya}\sur{Belykh}\hspace{2pt}\orcidlink{0009-0000-6983-1150}}
\author[1]{\fnm{Anastasia}\sur{Malakhova}\hspace{2pt}\orcidlink{0009-0007-7149-991X}}
\author[1]{\fnm{Alexander}\sur{Maloletov}\hspace{2pt}\orcidlink{0000-0002-7312-2944}}
\affil[1]{\orgname{Innopolis University}, \orgaddress{\city{Innopolis}, \country{Russia}}}

\abstract{We examine what plan validation and execution reports establish in
natural-language robot fleet control. Commands are traced through typed plans,
deterministic corrections, fleet coordination and outcome checks in a
three-level Robot Operating System 2 implementation with twenty simulated
robots. An offline benchmark of 44 cases checks plan structure and selected
command requirements. Across five local configurations without a response
schema, 70.0--87.3\% of responses pass. All runs in a selected 18-run simulation
series meet their pose-based task conditions. This series includes two
successful repeats that replaced incomplete runs after a software change. Saved
plans and logs document duplicate pickup assignments and goal adjustments within
the original task zones. Separate formation traces show a structurally accepted
plan with the wrong formation shape and a rejected route, and an arrival report
that omitted a requested robot after replanning. Carry completion records pickup
and subsequent drop-off zone presence for the same robot. The cases distinguish
changes to executable actions from changes to the requirements covered by a
completion report. They motivate checking the effective assignments, goals and
reported outcome against the original command.}

\maketitle

\section{Introduction}

\begin{figure}[!htbp]
\centering
\includegraphics[width=\textwidth]{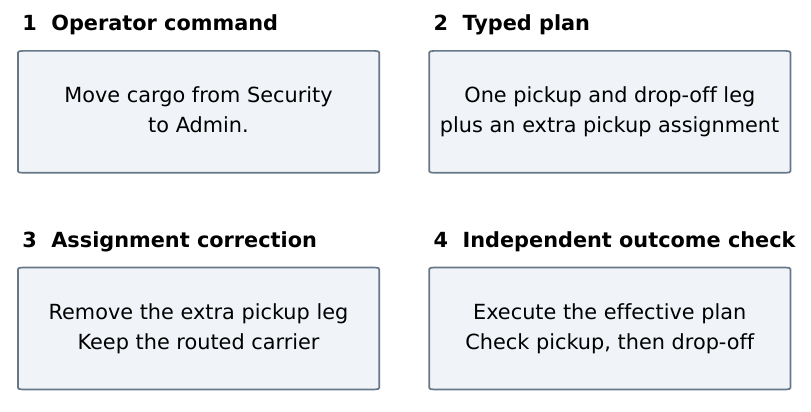}
\caption{Illustrative duplicate-carrier correction at the language--coordination
boundary. An additional robot is routed to the same pickup. The guard retains the
robot with the drop-off leg and removes the extra pickup assignment}

\label{fig:hero}
\end{figure}

Three recorded plans sent an extra robot to a pickup already assigned to
another carrier. Both assignments were valid under the plan format
(Fig.~\ref{fig:hero}). Language interfaces can express actions in terms of
skills and programs~\citep{saycan,codeaspolicies}. For a fleet, the
resulting assignments and goals still need to be coordinated across
robots sharing the same workspace.

We separate task interpretation from multi-agent path finding (MAPF).
Direct language-model generation of MAPF solutions has failed on tested
room and maze instances~\citep{mapfllmfail}. Our system instead sends
fleet-level actions to a dedicated planner.

Before a fleet moves, the system turns the operator's command into a plan,
checks or modifies its actions, and sends them to the coordination layer.
If a check changes an assignment or destination, we compare the action sent
to the fleet with the original command. We use an offline benchmark to study
plan validation and simulation logs to trace what happens when the system
changes an assignment or goal, rejects a formation route, or checks arrival
for a reduced robot set.

We ask which implemented validation criteria generated responses fail to satisfy
in the tested configurations (RQ1), which assignment and spatial conflicts the
checks handle in the recorded cases and how they change the effective plan
(RQ2), and which requirements of the original command are covered by a
component's success report after replanning (RQ3). The offline benchmark
addresses RQ1. The simulation plans, logs and task-state summaries address
RQ2 and RQ3.

\section{Related Work}

An instruction to a robot fleet leaves two different problems to solve:
deciding what the robots should do and finding motions that realize those
decisions. Classical multi-agent path finding addresses the latter once
starts, goals and movement rules have been specified. Conflict-Based
Search~\citep{cbs} resolves path conflicts through constraints, while
Priority-Based Search (PBS)~\citep{pbs} and large-neighborhood repair
(MAPF-LNS2)~\citep{lns2}
offer alternative ways to coordinate paths. The resulting paths
coordinate motion under the specified rules, but do not determine whether
the supplied assignments and goals express a
user's request. PlanBench documents failures of language models on its
own plan-generation tasks~\citep{planbench}.

Grounding language in executable skills is therefore central to early
robot planners based on large language models (LLMs). SayCan~\citep{saycan} combines a language model's
estimate of a skill's usefulness with an affordance estimate of whether the
robot can execute it in the current state. This connects task interpretation
to available behavior without asking the language model to generate motor
commands. Code as Policies~\citep{codeaspolicies} gives the model a more
expressive interface: it composes perception and control APIs into programs,
including control flow and numerical operations. For fleet control the
distinction matters: choosing an available action and assembling a coherent
mission require different checks.

Execution can depart from a plan even when its actions are available. Inner
Monologue~\citep{innermono} feeds observations and success reports back into
language reasoning, allowing subsequent decisions to account for execution.
ProgPrompt~\citep{progprompt} incorporates this concern into program-like
plans through assertions and recovery steps. In its simulated evaluation,
assertions are assessed by an LLM using environment-state information,
whereas our runtime guards apply explicit predicates. In both systems,
feedback informs the next planning decision.

For a robot team, grounding individual actions must be combined with
allocation and coordination~\citep{mrssurvey}.
The task-allocation taxonomy distinguishes single- and multi-task robots,
single- and multi-robot tasks, and instantaneous and time-extended
assignment~\citep{mrta}. Our duplicate-carrier case concerns the assignment
of a single-robot task. SMART-LLM~\citep{smartllm}
uses prompted task decomposition, coalition formation and allocation to
construct a team plan. The language model takes part in selecting the robots.
RoCo~\citep{roco} uses dialogue between robot-specific language agents to
develop a joint plan, which is checked before a centralized multi-arm planner
generates motion. Our duplicate-carrier and shared-destination cases examine
how proposed assignments and goals are handled during execution.

LaMMA-P and HiP-CL construct explicit planning problems for robot
teams~\citep{lammap,hipcl}. LLM+P translates instructions into the Planning
Domain Definition Language (PDDL) and delegates search to a classical
planner~\citep{llmp}. LLM-Modulo uses external checks to evaluate model
proposals~\citep{llmmodulo}. Temporal-logic synthesis can give guarantees
relative to a realizable specification and an environment
model~\citep{ltlmission}.

CommandSwarm validates generated behavior trees against an executable set of
swarm primitives, explicitly distinguishes parser acceptance from semantic
correctness, and discusses logging commands, generated trees and execution
status for auditability~\citep{commandswarm}. Sangchai et
al.~\citep{symbolicwrappers} combine language reasoning with symbolic wrappers
in a three-level architecture. Their tool-based and PDDL-based variants expose
intermediate plans for human inspection and revision. Khatana et
al.~\citep{verifiermediated} ground candidate multi-robot plans into
trajectories and return geometric conflict information to the planners.
We use execution logs to see which parts of the original command survive
runtime adjustments and which parts an arrival report covers.

\section{System Overview}
\label{sec:overview}

\begin{figure}[!htbp]
\centering
\includegraphics[width=0.64\textwidth]{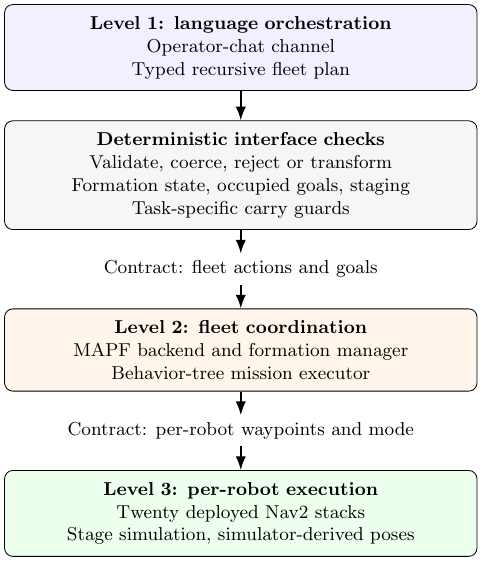}
\caption{Operator-chat path from a typed fleet plan through interface checks,
fleet coordination, and per-robot execution}
\label{fig:arch}
\end{figure}

Figure~\ref{fig:arch} shows the operator-chat path. Level~1 turns a natural
language command into a typed plan and applies interface checks. Level~2
coordinates the resulting fleet actions. Level~3 tracks the scheduled
waypoints and executes the control commands.

The interface exposes four distinct correctness questions. Structural
validity concerns admissible node types and fields. Task meaning concerns
the assignments and goals implied by the operator's command. Geometric
feasibility concerns whether robots or formations can traverse the required
space. Evaluator completion concerns whether the specified pose-based events
occur. No implication between these properties follows from the typed
representation alone. Each result must therefore be interpreted with
respect to the plan, robot set, goals or task conditions to which the
corresponding check was applied. Table~\ref{tab:coverage} links each check to the
corresponding evidence.

\begin{table}[!htbp]
\caption{Implemented checks and their observed scope}
\label{tab:coverage}
\centering
\setlength{\tabcolsep}{4pt}
\begin{tabular}{>{\raggedright\arraybackslash}p{0.21\textwidth}
                >{\raggedright\arraybackslash}p{0.33\textwidth}
                >{\raggedright\arraybackslash}p{0.34\textwidth}}
\toprule
Property & Check in this implementation & Evidence and limit \\
\midrule
Plan representation & Recursive node and field checks, with selected type coercions
 & E1 also checks case-specific requirements. Acceptance leaves other mission properties open. \\
Robot assignments & Active-formation membership and single-carrier rules
 & Single-carrier rule fired in two selected M2 runs. \\
Occupied goals & Chat-level goal adjustment and planner-level blocked-cell substitution
 & Both mechanisms appear in M4 run \code{20260818\_140918}. No chat rewrites in the 13 selected language runs. \\
Shared drop-off & Spread coincident goals for distinct carry tasks
 & Logged in three M3 runs. The design does not isolate its effect. \\
Formation assembly & Stage followers and check pose tolerances
 & M4 includes an assembled formation and a separate staging failure. \\
Formation route & Static path check with the planner's footprint approximation
 & LNS2 rejected the leader's route in M4 run \code{20260819\_073009}. No pre-assembly check of the full oriented formation route. \\
Task completion & Pose samples in specified zones at 5\,Hz, with pickup before drop-off
 & All 18 selected M2/M3 runs satisfy this evaluator. Physical transport and intent are outside the metric. \\
Repair and supervision & Execution remediation, verification repair and bounded mission supervision
 & One supervision pass per selected language run. No remediation or verification-repair attempts in that series. \\
\bottomrule
\end{tabular}
\end{table}

Twenty differential-drive robots operate in Stage~\citep{stage} on the
\code{amongus} map ($66.6 \times 37.75$\,m, Fig.~\ref{fig:map}). The world
file places five groups of four around their respective home locations.
The number deployed is distinct from the number moving in any interval.

\section{Level 1: Language Orchestration}
\label{sec:orch}

In the operator-chat channel, a response contains a natural-language reply
and a plan tree in a JavaScript Object Notation (JSON) object. Four leaf types cover fleet actions --- \code{mapf} (move a robot set to
a goal set), \code{formation} (assemble and hold), \code{disband}, and
\code{idle} (returns no action and can carry an escalation reason). Two
containers, \code{sequence} and \code{parallel}, nest recursively. This
example illustrates the plan format:

\begin{center}\small
\begin{minipage}{\columnwidth}
\begin{verbatim}
{"type":"sequence","steps":[
 {"type":"mapf","robot_ids":[8,9,10,11],
  "goals":[[-8.7,-5.9],[-9.6,-5.9],
           [-8.7,-6.8],[-9.6,-6.8]]},
 {"type":"formation","formation_id":"column_8",
  "leader_ns":"robot_8",
  "follower_ns":["robot_9",
   "robot_10","robot_11"],
  "offsets_x":[-1.5,-3.0,-4.5],
  "offsets_y":[0,0,0]}]}
\end{verbatim}
\end{minipage}
\end{center}

The JSON schema and recursive validator check the structure of the plan.
Checking task requirements and route feasibility requires additional information.

A \code{sequence} executes its children in order and terminates on the first
failure. For a \code{parallel} container, the executor first merges compatible
MAPF leaves into a single fleet-level request. Non-MAPF actions follow, then
nested sequences in their original order. Each dispatched command completes
before the next begins. If the executor encounters an \code{idle} leaf during
this transformation, it reduces the subtree to that idle action.

The system has two other language channels. The reactive channel returns a
wait, abort, or replan decision to a behavior-tree node on warning or error.
The proactive observer can send a
flat mode command to the behavior tree. The operator-chat channel uses the
recursive plan grammar and the guards described here. E1 exercises offline
generation for that grammar, and the language runs in E2 use the
operator-chat path. That path can query live pose and task state and compute
room placements through eight allowlisted tools.

E1 compares text generation with and without a response schema. Both modes
parse the returned plan and apply case validators. In E2, the operator chat
can make auxiliary tool calls before returning the plan. Both E1 and E2
receive the plan as JSON.

The operator-chat path validates node types and fields, coerces selected
values, guards active-formation membership, adjusts occupied goals, and
stages robots before assembly. The single-carrier rule removes a redundant
pickup assignment when it can identify the robot routed onward to the
drop-off. The shared-drop-off rule spreads coincident destinations in
eligible cases. The task evaluator monitors the original zones during
execution, including when a guard changes a waypoint.

The operator chat can attempt remediation after an execution error or repair
a plan after checking robot poses and formation status. Mission supervision
can request additional work within configured limits.

\section{Level 2: Fleet Coordination}

The PBS and LNS2 implementations accept robot identifiers and a goal for
each robot through the same action interface. They publish scheduled
waypoints to the robots. The simulation runs reported here use LNS2, which
checks static reachability with its footprint model before planning.

A formation specification names a leader, followers and per-follower
offsets in the leader's frame. The coordination layer excludes active
followers from independent MAPF assignment and represents the group through
the leader's enlarged planning footprint. This approximation is distinct
from planning with the full, orientation-dependent group shape.
BehaviorTree.CPP~\citep{btcpp} runs the mission tree.

The task evaluator samples robot positions during execution at 5\,Hz. A visit
task completes when a pose sample shows a robot within 1.5\,m of the target.
For a carry task, the evaluator records the first robot it detects in the
pickup zone and marks the task complete when it later detects that same
robot in the drop-off zone. Each transition requires a single in-zone
sample. The evaluator records zone presence independently of component
status reports. Physical load handling is outside the metric.

\section{Level 3: Per-Robot Execution}
\label{sec:lvl3}

Each robot runs a namespaced Robot Operating System 2 (ROS~2) Humble~\citep{ros2}
Nav2~\citep{nav2} stack with staggered startup. Stage supplies the poses
used for navigation and task evaluation. Adaptive Monte Carlo localization
(AMCL) is disabled.

\section{Experimental Design}
\label{sec:design}

The analysis uses the saved benchmark outputs, plans, logs and derived
metrics in the repository, under \code{paper/results}.

\subsection{E1: Offline plan validation}
The benchmark has 44 cases in twelve categories covering navigation,
parallel and sequential composition, formation, disbanding, idle,
escalation and edge cases. Each reported model and decoding condition
contains five scored attempts per case at temperature 0.1. The parser
extracts the plan and applies structural checks. Case validators then
test selected requirements such as robot identifiers, proximity to a
specified destination, and step ordering. A pass means that the response
satisfies those checks. No robots are run in E1.

For example, case \code{mapf\_explicit\_01} checks the required robot
identifiers and the number and distinctness of goals. It leaves the requested
coordinates unchecked. Case \code{sequence\_04} checks the number of steps and
the presence of an idle node without checking the requested navigation steps.

Schema-off requests use unconstrained text generation and parse a JSON
plan from the result. Schema-on requests pass the plan schema as the
response format. Neither offline condition submits tool definitions.
The local sweep used Ollama, a 16\,GB RTX~5060~Ti, a 32{,}768-token context,
and a 2{,}048-token output limit. The local manifest records partial CPU
offload for Mistral Small 3.2. Hosted runs used OpenRouter with an
8{,}192-token output limit. These are comparisons of the logged
configurations, which differ in model, serving stack, and sometimes
provider or token budget.

Hosted transport errors were handled by rerunning all five attempts of the
affected case and replacing that case's original records. The raw records
log replacements for 18 schema-off and 44 schema-on Nemotron cases, nine
cases in each 397B condition, and 17 schema-on hosted 9B cases. These cells
can mix provider routes. Thus 220 is the number of scored records per cell.
The harness, its prompt builders, per-case prompts, responses, and scoring
scripts support reanalysis of the saved outputs. Exact model snapshots and
complete per-request serving configurations are unavailable.

\subsection{E2: Task completion in simulation}
\begin{figure}[!htbp]
\centering
\includegraphics[width=\textwidth]{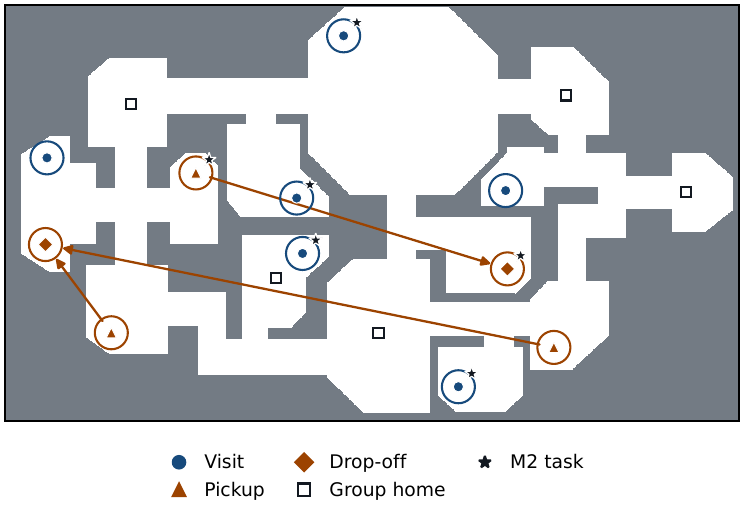}
\caption{The \code{amongus} map ($66.6 \times 37.75$\,m, 0.05\,m cells) and
the eleven distinct zones of radius 1.5\,m used by the nine M3 tasks. Circles show the
zone boundaries and center markers identify visit, pickup, and drop-off
zones. Arrows link pickup and drop-off locations. Stars mark the M2 subset.
Squares mark the home locations of the five groups of four robots}
\label{fig:map}
\end{figure}

M2 consists of four visit tasks and one carry task. M3 consists of six visit
tasks and three carry tasks, two of which share a drop-off location
(Fig.~\ref{fig:map}). The selected descriptive series contains five scripted
M2 runs with fixed assignments, five M2 runs using hosted Qwen3.5-9B, five
M2 runs using hosted Qwen3.5-397B-A17B (cell B), and three M3 runs using the
397B model (cell B-M3). The scripted driver uses the same coordination and
execution layers. Model, routing, allocation and software version vary
across the reported cells.

Selection depended on outcome in B/M2. The
\href{https://github.com/Innopolis-Robotics-Society/llm_swarm}{development record} (\code{paper/results/e2\_mission\_series})
reports
that three original runs completed all five tasks, while runs
\code{20260818\_085836} and \code{20260818\_090145} each stalled at 4/5.
After the single-carrier guard was introduced, those two runs were
replaced by \code{20260818\_093216} and \code{20260818\_093704}. The selected
B/M2 cell thus combines three original successes and two successful
repeats. The bags of the two superseded runs are absent from the public
archive, so their scores are taken from the run history. The 18-run
analysis estimates neither a campaign-wide completion rate nor a guard
effect. Table~\ref{tab:provenance} summarizes the run selection and the
available evidence for each series.

\begin{table}[!htbp]
\caption{Provenance and analysis populations for the simulation evidence}
\label{tab:provenance}
\centering
\setlength{\tabcolsep}{4pt}
\begin{tabular}{>{\raggedright\arraybackslash}p{0.20\textwidth}
                >{\raggedright\arraybackslash}p{0.14\textwidth}
                >{\raggedright\arraybackslash}p{0.54\textwidth}}
\toprule
Series & Runs & Inclusion and evidence \\
\midrule
Selected M2/M3 & 18 & Four cells. Session IDs and code-version
 labels are recorded. All 18 worktrees were dirty. Thirteen runs have a language
 journal, of which eight have per-phase checkpoints. \\
Superseded B/M2 & 2 & Two runs at 4/5 tasks replaced after a software change.
 The three original successes remain in the selected cell. \\
M3 before/after & 2 additional & The pre-guard runs have logs and derived
 metrics but no decision journal. The three post-guard runs are already
 counted in the B-M3 cell above. \\
M4 formation & 4 recorded attempts & Two retained qualitative cases
 (\code{20260818\_140918}, \code{20260819\_073009}). A pilot
 (\code{20260818\_135527}) stopped early and a 42\,s recording
 (\code{20260819\_072231}) with no motion was discarded. M4 is outside
 the task-completion scoring for M2/M3. \\
\bottomrule
\end{tabular}
\end{table}

The archived session files and logs identify the included runs. A recorded
code version does not identify the uncommitted changes present in a run. The
comparison uses the task evaluator recorded in \code{/tasks/state} and
the target task list defined for each mission. For completed runs, the
task-completion interval runs from the first recorded task-state sample to
the first sample in which the last target is marked complete, using
simulation time. Language-loop duration uses the final cumulative
wall-clock counter in each language run's mission journal.

\Needspace{12\baselineskip}
\section{Results}
\label{sec:results}

\subsection{Assignment defects and shared-destination conflicts}
\label{sec:absorb}

Three of the 13 selected language runs have a saved plan that sends an
extra robot to the pickup. In two further runs, the single-carrier guard
counter records a correction and the saved plan already reflects it.
All three model-9b runs with a visible extra assignment completed M2.
Both B/M2 replacements also completed M2 after the guard removed an
extra pickup assignment.

M3 assigns two carry tasks to the same drop-off center
(Fig.~\ref{fig:map}), so both carriers can be directed to the same point.
Both archived pre-guard runs (\code{20260815\_153035} and
\code{20260815\_154541}) complete 8/9 tasks. Each of the three B-M3 runs
completes 9/9. In run \code{20260818\_111548} the guard changed the common
target $(-29.65,-2.85)$\,m to $(-28.90,-2.85)$\,m for robot~3 and
$(-30.40,-2.85)$\,m for robot~16. Both goals are 0.75\,m from the specified
center and inside the original evaluator zone. The logs show spreading
in all three post-guard runs. The two series differ in software version
and provider routing, and the pre-guard journals are absent. These records
document the intervention and outcomes without isolating its effect.

\subsection{Case study: an assembled formation with a rejected route}
The M4 session
\href{https://github.com/Innopolis-Robotics-Society/llm_swarm}{\code{20260819\_073009}}
was interactive. The initial column request produced longitudinal offsets
\code{offsets\_x=[-1.5,-3.0,-4.5]} and \code{offsets\_y=[0,0,0]}
(mission \code{83f336e2bf07}). After the operator requested that robot~0 be
brought to the required position, a repeated column request produced
\code{offsets\_x=[0,0,0]} and \code{offsets\_y=[1.5,-1.5,3.0]}
(mission \code{c16217cb2ab3}), describing a line abreast with a nominal span
of 4.5\,m between the outer centers. The log records activation with those
offsets. On the following move, LNS2 reported no static path for the leader
with the footprint approximation on the planning grid used for that request.
The session-level derived status metrics report 6.5\,s from the first active
sample to the first stable sample and no broken samples while active.

The latter plan passed structural validation and activated a formation whose
shape differed from the request. The route was then rejected under the
planner's approximation. These observations belong to different command
episodes within the same session. In M4 run \code{20260818\_140918}, the
recorded plan also used column offsets \code{offsets\_x=[-1.5,-3.0,-4.5]}.

\subsection{Local success reports and the original request}
\label{sec:failures}

In M4 run
\href{https://github.com/Innopolis-Robotics-Society/llm_swarm}{\code{20260818\_140918}},
MAPF reported \emph{All agents arrived}
while robot~0, part of the original request, remained at its starting
position. The development record traces the mismatch to a replan that
removed robot~0 from the set used to check arrival.

In a later step of the same run, the planner substituted a free cell for
a blocked goal. The formation manager reported robot~3 at 1.97\,m from
its required slot, outside the 0.50\,m tolerance. The executor learned
of the refusal at its step timeout.

\subsection{Offline plan validation results}
\label{sec:planacc}

Table~\ref{tab:e1} reports a response-level pass rate and the share of
cases with at least one pass in five attempts. Table~\ref{tab:e1errors}
breaks down the 259 failed responses among the 1{,}100 local schema-off
responses. For Mistral Small 3.2, the same offsets-arity message accounts
for 20 of 47 failures.

For the hosted Nemotron schema-off condition, reapplying the current case
validators to the saved responses changes two \code{edge\_06} outcomes from fail
to pass: both use the accepted \emph{clarify:} prefix. Table~\ref{tab:e1}
reports the rescored values, with 201 passing responses out of 220 and at least
one pass in all 44 cases.

\begin{table}[!htbp]
\centering
\caption{Offline validation on 44 cases with five scored responses per case
and condition. Call rate is the percentage of 220 responses that passed.
Case rate is the percentage of 44 cases with at least one pass in five
attempts. Model sizes are nominal. For mixture-of-experts models they are
total/active parameters. n/d denotes an undisclosed size. A dash denotes an
unreported condition}
\label{tab:e1}
\setlength{\tabcolsep}{3.2pt}
\begin{tabular}{lccccc}
\toprule
& & \multicolumn{2}{c}{Schema off} & \multicolumn{2}{c}{Schema on} \\
\cmidrule(lr){3-4}\cmidrule(lr){5-6}
Model & Size & Call (\%) & Case (\%) & Call (\%) & Case (\%) \\
\midrule
\multicolumn{6}{l}{\emph{Local}}\\
Qwen3.5-4B & 4B & 70.0 & 93.2 & 77.7 & 95.5 \\
Gemma 3 12B & 12B & 71.4 & 72.7 & --- & --- \\
Qwen2.5-14B & 14B & 75.0 & 81.8 & 75.5 & 81.8 \\
Qwen3.5-9B & 9B & 87.3 & 95.5 & 19.1 & 22.7 \\
Mistral Small 3.2 & 24B & 78.6 & 81.8 & --- & --- \\
\midrule
\multicolumn{6}{l}{\emph{Hosted}}\\
Qwen3.5-9B & 9B & 90.5 & 97.7 & 51.4 & 61.4 \\
Gemini 2.5 Flash & n/d & 89.5 & 90.9 & 80.9 & 86.4 \\
Nemotron 3 Ultra & 550B/55B & 91.4 & 100.0 & 88.6 & 95.5 \\
Qwen3.5-397B-A17B & 397B/17B & 93.2 & 100.0 & 95.0 & 100.0 \\
\bottomrule
\end{tabular}
\end{table}

\begin{table}[!htbp]
\centering
\caption{Failure categories for local schema-off responses in E1. The
scoring script assigns one category per failed response using a fixed
priority order over error messages}
\label{tab:e1errors}
\begin{tabular}{lr}
\toprule
Category & Failed responses \\
\midrule
Formation fields or staging & 82 \\
Plan structure & 81 \\
Goal checks & 34 \\
Unparseable JSON & 34 \\
Other validator messages & 22 \\
Escalation reason & 5 \\
Robot identifiers & 1 \\
\midrule
Total & 259 \\
\bottomrule
\end{tabular}
\end{table}

\subsubsection{Decoding configurations}
\label{sec:hosted}
When a response schema was requested, the local 9B score was 19.1\%,
versus 87.3\% without one. The local 4B score was 77.7\%, versus
70.0\%, and the local 14B condition gained one passing response out of
220. Hosted differences also vary in direction. The 397B/17B scores were
95.0\% with a schema and 93.2\% without one, while Gemini scored 80.9\%
and 89.5\%, respectively. These comparisons describe the logged
configurations.

\subsection{Task-evaluator completion in the selected series}
All 18 included runs satisfy their target task conditions
(Table~\ref{tab:e2}). The B/M2 row combines the three successful original
runs with the two successful replacements.

\begin{table}[!htbp]
\centering
\caption{Selected M2/M3 runs. Task counts are pose-based outcomes per run.
The task interval uses simulation seconds and is reported as mean and sample
standard deviation (SD). Language-loop time is the mean wall-clock duration,
including auxiliary calls and retries. The version log reports that 3/5
original B/M2 runs completed all tasks. The two runs that stopped at 4/5
tasks were replaced after a software change}
\label{tab:e2}
\setlength{\tabcolsep}{4pt}
\begin{tabular}{llcccc}
\toprule
Cell & Mission & \makecell{Full\\runs} & \makecell{Tasks\\per run}
 & \makecell{Task interval (sim s)\\mean $\pm$ SD} & \makecell{Language\\loop (wall s)} \\
\midrule
no-LLM   & M2 & 5/5 & 5/5 & $203.5 \pm 10.6$ & 0 \\
model-9b & M2 & 5/5 & 5/5 & $252.2 \pm 26.6$ & 9.5 \\
B        & M2 & 5/5 & 5/5 & $260.0 \pm 16.5$ & 31.6 \\
B-M3     & M3 & 3/3 & 9/9 & $587.7 \pm 67.0$ & 51.2 \\
\bottomrule
\end{tabular}
\end{table}

The M2 mean task interval ranges from 203.5 to 260.0 simulation seconds.
The mean language-loop duration ranges from 0 to 31.6 wall-clock seconds.
These intervals start at different events and use different clocks.
Recorded odometry summaries show that 9--10 distinct robots per selected
run traveled at least 0.5\,m during the recording.

None of the 13 selected language runs invoked remediation or verification
repair, and none triggered the active-formation guard or chat-level
occupancy rewrites. Each logged one supervision step. Twelve runs made no
auxiliary calls. One B-M3 run (\code{20260818\_113030}) called
\code{list\_tasks} once and \code{verify\_plan\_execution\_state} twice.
The two additional M4 records contain 10 and 33 auxiliary calls.

\section{Discussion}
\label{sec:discussion}

The E1 pass rates compare generation settings against the parser and the
selected case checks. Since the benchmark has no paired execution outcomes,
it cannot tell us how often a passing response would complete the mission.

The two carry rules change different parts of the plan. The single-carrier
rule changes which robot is assigned. The shared-drop-off rule changes
waypoints within the original task zone, so the adjusted goals can still
satisfy the evaluator. In the formation case, the robots assembled to the
generated offsets even though those offsets described a different shape from
the one requested. The planner then rejected the route under its footprint
approximation.

The M4 arrival report applied to the robot set remaining after replanning.
Comparing that set with the original request exposed the omitted robot.
The final chat record marked the mission unsuccessful. The inconsistent
success message came from the MAPF component.

These observations support reporting the requested, planned, skipped and arrived
robot sets separately, together with effective goals and any task regions that
permit goal adjustment. When the original command requires a particular robot to
reach a goal, removing that robot from a planning request leaves the requirement
unresolved unless it is completed or explicitly revised. The cases motivate
this reporting practice. They leave the causal effect of the guards open.

\section{Limitations}
\label{sec:limitations}

The simulation evidence covers one Stage map, two main missions and
selected runs. The version log documents the B/M2 replacements. The M3
before/after series differ in software version and provider routing, so
the comparison cannot isolate the guard's effect. The recorded code
versions do not capture uncommitted changes in the historical runs.

E1 repeats 44 case designs and provides a comparison of the logged model
and serving configurations. Exact model revisions and complete serving
settings are unavailable. The simulation task metric records zone
presence. It does not measure physical cargo transfer, collision safety
or peak concurrency. The M4 observations illustrate individual failures
and do not estimate their frequency.

\section{Conclusion}

This case study traced natural-language fleet commands through typed plans,
runtime corrections, motion coordination and completion reports. The recorded
cases show how these stages can refer to different assignments, destinations and
robot sets: an accepted plan included an extra pickup assignment, a formation
assembled to offsets that differed from the requested shape, and an arrival
report covered a reduced robot set after replanning.

The practical implication is to associate each validation result and success
report with the requirements it covers. Assignment and waypoint changes should
remain traceable to the original command, including task regions that allow
alternative destinations. Completion reporting should retain visibility of
requested robots omitted during planning. This makes the scope of a local
success explicit when assessing the fleet's response to the operator.

\backmatter

\section*{Statements and Declarations}

\subsection*{Competing Interests}
The authors have no relevant financial or non-financial interests to disclose.

\subsection*{Funding}
This work was supported by internal funding from Innopolis University.

\subsection*{Ethics Approval}
Not applicable. This study did not involve human participants or animals.

\subsection*{Consent to Participate}
Not applicable.

\subsection*{Consent for Publication}
Not applicable.

\subsection*{Use of Generative AI}
OpenAI Codex (0.135.0) and Anthropic Claude Code (2.1.159) were used to write
the implementation code and as assistants in preparing the manuscript,
including the figures. All text, results and interpretations were reviewed by
the authors, who take full responsibility for the content.

\subsection*{Availability of Data and Material}
The benchmark outputs, session metadata, logs, derived metrics, and
analysis scripts are in the
\href{https://github.com/Innopolis-Robotics-Society/llm_swarm}{repository}, under \code{paper/results}.
The recorded bags are available in the
\href{https://drive.google.com/drive/folders/1fpkuhE1f3ET2YIJLI5sV4jPzE1IJophs?usp=sharing}{public results archive},
\code{llm\_swarm\_results\_2026-11-04.tar.zst} (135\,MB).
Its SHA-256 checksum is
\code{3866f1fce6254243\hspace{0pt}986b0052e2fc0d60\hspace{0pt}39eadc351e10c460\hspace{0pt}f4ed0feec69f740d}.
A per-file manifest accompanies the archive. With the archive unpacked,
run the repository's \code{paper/results/scripts/analyze\_sessions.py}
with \code{--sessions} pointing to the unpacked
\code{results/e2\_mission\_series} and pass \code{--json e2\_metrics.json}
to export per-run metrics.
The two superseded B/M2 bags are absent. Their scores and replacement
identifiers are documented in the run history.

\subsection*{Code Availability}
The ROS~2 stack, benchmark harness, and analysis scripts are available at
\href{https://github.com/Innopolis-Robotics-Society/llm_swarm/}{repository}.
This revision records the analysis and documentation. It does not
capture the uncommitted changes in the historical simulation runs.
Licence declarations, where present, are in the individual packages.

\subsection*{Authors' Contributions}
A.T.: Conceptualization, Methodology, Software, Investigation, Data curation,
Writing---original draft, Writing---review and editing, Project administration.
D.V.: Software, Investigation, Writing---review and editing.
E.S.: Conceptualization, Methodology, Software, Investigation, Supervision,
Project administration, Writing---review and editing.
S.B.: Software, Investigation, Visualization.
A.Mal.: Software, Validation.
A.Malol.: Supervision, Conceptualization, Writing---review and editing.

\subsection*{Acknowledgements}
Not applicable.


\begingroup
\let\section\unpatchedsection
\setlength{\bibsep}{0.5em}
\interlinepenalty=10000
\bibliography{refs}
\endgroup

\clearpage
\section*{Author Biographies}

\begin{biography}{}{\author{Artem Tuzov} is an undergraduate student at
Innopolis University specializing in robotics. His work focuses on robot
perception and language-model interfaces for multi-robot systems.}
\end{biography}

\begin{biography}{}{\author{Dmitrii Vizitei} is an undergraduate student in
data science and artificial intelligence at Innopolis University. He works on
ROS~2 software and the integration of language models with robot control.}
\end{biography}

\begin{biography}{}{\author{Evgenii Shlomov} is an undergraduate student at
Innopolis University and leads the Innopolis Robotics Society. His work
includes motion planning for industrial manipulators, autonomous mobile
navigation, and multi-robot coordination.}
\end{biography}

\begin{biography}{}{\author{Senya Belykh} is an undergraduate student at
Innopolis University. He works on formation control for mobile robots,
including leader--follower coordination and geometric footprint modeling.}
\end{biography}

\begin{biography}{}{\author{Anastasia Malakhova} is an undergraduate student
in data science and artificial intelligence at Innopolis University. Her work
includes computer vision, object detection, and software for robotic systems.}
\end{biography}

\begin{biography}{}{\author{Alexander Maloletov} is a professor and director
of the Institute of Robotics and Computer Vision at Innopolis University.
He holds a Doctor of Sciences degree in physics and mathematics and works
on theoretical mechanics and robotics.}
\end{biography}

\end{document}